\documentclass[conference]{IEEEtran}
\IEEEoverridecommandlockouts
\usepackage{cite}
\usepackage{amsmath, epsfig, ulem, amssymb, bm, bbm}
\usepackage{hyperref}
\usepackage{subfigure}
\usepackage{booktabs} 
\usepackage{bbding}
\usepackage{graphicx}
\usepackage{float}
\usepackage{array}
\usepackage{caption}
\usepackage{multirow}
\usepackage{enumitem}
\usepackage{multicol}
\usepackage{tabularx}
\usepackage{threeparttable}
\usepackage{verbatim}
\usepackage{bbding}

\usepackage{xcolor}
\usepackage{stfloats}
\usepackage[linesnumbered,ruled,vlined]{algorithm2e}
\SetKwInput{KwInput}{Input}                
\SetKwInput{KwOutput}{Output}              

\usepackage{amsmath,amssymb,amsfonts}
\usepackage{algorithmic}
\usepackage{graphicx}
\usepackage{textcomp}
\usepackage{xcolor}
\def\BibTeX{{\rm B\kern-.05em{\sc i\kern-.025em b}\kern-.08em
    T\kern-.1667em\lower.7ex\hbox{E}\kern-.125emX}}
\begin{document}

\title{AU-aware graph convolutional network for Macro- and Micro-expression spotting
}

\author{\IEEEauthorblockN{Shukang Yin}
\IEEEauthorblockA{University of Science and \\
Technology of China \\
Hefei, Anhui, China \\
xjtupanda@mail.ustc.edu.cn
}\\   
\IEEEauthorblockN{Shifeng Liu}
\IEEEauthorblockA{University of Science and \\
Technology of China \\
Hefei, Anhui, China \\
lsf0619@mail.ustc.edu.cn
}
\and
\IEEEauthorblockN{Shiwei Wu}
\IEEEauthorblockA{University of Science and \\
Technology of China \\
Hefei, Anhui, China \\
dwustc@mail.ustc.edu.cn
}
 \\ 
\IEEEauthorblockN{Sirui Zhao\IEEEauthorrefmark{1}}
\IEEEauthorblockA{University of Science and \\
Technology of China \\
Hefei, Anhui, China \\
sirui@mail.ustc.edu.cn
}
\and
\IEEEauthorblockN{Tong Xu}
\IEEEauthorblockA{University of Science and \\
Technology of China \\
Hefei, Anhui, China \\
tongxu@ustc.edu.cn
} \\ 
\IEEEauthorblockN{%
Enhong Chen\IEEEauthorrefmark{1}\thanks{\IEEEauthorrefmark{1}Sirui Zhao and Enhong Chen are corresponding authors.}}
\IEEEauthorblockA{University of Science and \\
Technology of China \\
Hefei, Anhui, China \\
cheneh@ustc.edu.cn
}

}

\maketitle

\begin{abstract}
Automatic Micro-Expression (ME) spotting in long videos is a crucial step in ME analysis but also a challenging task due to the short duration and low intensity of MEs. When solving this problem, previous works generally lack in considering the structures of human faces and the correspondence between expressions and relevant facial muscles. To address this issue for better performance of ME spotting, this paper seeks to extract finer spatial features by modeling the relationships between facial Regions of Interest (ROIs). Specifically, we propose a graph convolutional-based network, called Action-Unit-aWare Graph Convolutional Network (AUW-GCN). Furthermore, to inject prior information and to cope with the problem of small datasets, AU-related statistics are encoded into the network. Comprehensive experiments show that our results outperform baseline methods consistently and achieve new SOTA performance in two benchmark datasets, $\text{CAS(ME)}^2$ and SAMM-LV. Our code is available at \href{https://github.com/xjtupanda/AUW-GCN}{https://github.com/xjtupanda/AUW-GCN}.
\end{abstract}


\section{Introduction}
\label{sec:intro}

Facial expressions highly reflect people's emotions and convey their psychological states in a non-verbal form~\cite{liong2022mtsn}. According to intensity and duration, they can be divided into two categories, including Macro-Expression (MaE) and Micro-Expression (ME). Generally, MaEs are more intense and enduring than MEs, usually lasting from 0.5 to 4.0s, and can be faked with intention. In contrast, MEs appear subtle, and their duration is generally less than 0.5s~\cite{wang2021mesnet}. Moreover, MEs are involuntary, spontaneously manifesting themselves when people try to hide their genuine feelings~\cite{ekman2009telling}. In recent years, MEs have been brought to the attention of the research community due to their wide applications, such as criminal investigations~\cite{ekman2009telling}, clinical psychology~\cite{salter2005sex} and national security~\cite{oh2018survey}. Specifically, an essential part of ME research is ME spotting, which involves locating expression intervals in long untrimmed videos. However, the task is quite tricky in itself due to the characteristics of ME, i.e., low intensity and short~duration. 

In recent years, much progress has been made in ME spotting. Previous studies mainly used raw features as input, including RGB images and optical-flow maps. Yap et al.~\cite{yap20223d} designed RGB sample pairs as model inputs for ME and MaE respectively. However, RGB images are insufficient in characterizing facial movements, especially those of MEs. Therefore, later works such as SOFTNet~\cite{liong2021shallow} exploited optical-flow features to describe the subtle motion of facial muscles, and ABPN~\cite{leng2022abpn} used finer MDMO~\cite{liu2015main} features. Nevertheless, by inputting the pre-processed features as a whole into the network, these works did not fully consider the relationships between different parts of human faces as well as the semantic information of these regions. However, this information is crucial to spotting performance because facial expressions appear through the movement of corresponding facial muscle groups, termed Action Units (AUs)~\cite{rosenberg2020face}, and ignorance of this semantic information could bring about confusion and ambiguity in the model. Therefore, this paper proposes to model the relationships between facial ROIs and leverage information of AUs to help extract finer feature representation and boost spotting performance.

To this end, we put forward an AU-aware graph convolutional network (GCN) for spotting MaEs and MEs in long untrimmed videos, dubbed AUW-GCN. Specifically, our model adopts a GCN~\cite{kipf2016semi} as the feature embedding backbone to characterize the relationships between different regions on human faces. After that, we modify an ABPN~\cite{leng2022abpn} for temporal feature interaction to better utilize contextual information in the videos. Moreover, in an effort to inject prior information about facial expressions and mitigate the issue of over-fitting brought by small datasets, we propose to encode AU statistics into the GCN to capture motion patterns of facial expressions better and achieve finer feature representation. Our main contributions can 
be summarized as follow:

\begin{itemize}[noitemsep,nolistsep]
    \item We propose a model to capture finer spatial features, thus making expression spotting more accurate and~complete.
    \item We design a strategy to encode the prior information about motion patterns of facial expressions into our network, which is crucial to refining spatial feature embedding and alleviating over-fitting.
    \looseness=-1
    \item We demonstrate the effectiveness of our proposed approach through comprehensive experiments on two standard benchmark datasets, $\text{CAS(ME)}^2$ and SAMM-LV.
\end{itemize}

\begin{figure*}[!th]
\begin{center}
\begin{minipage}[b]{\textwidth}
  \centering
  \centerline{\includegraphics[width=0.98\textwidth,height=200pt]{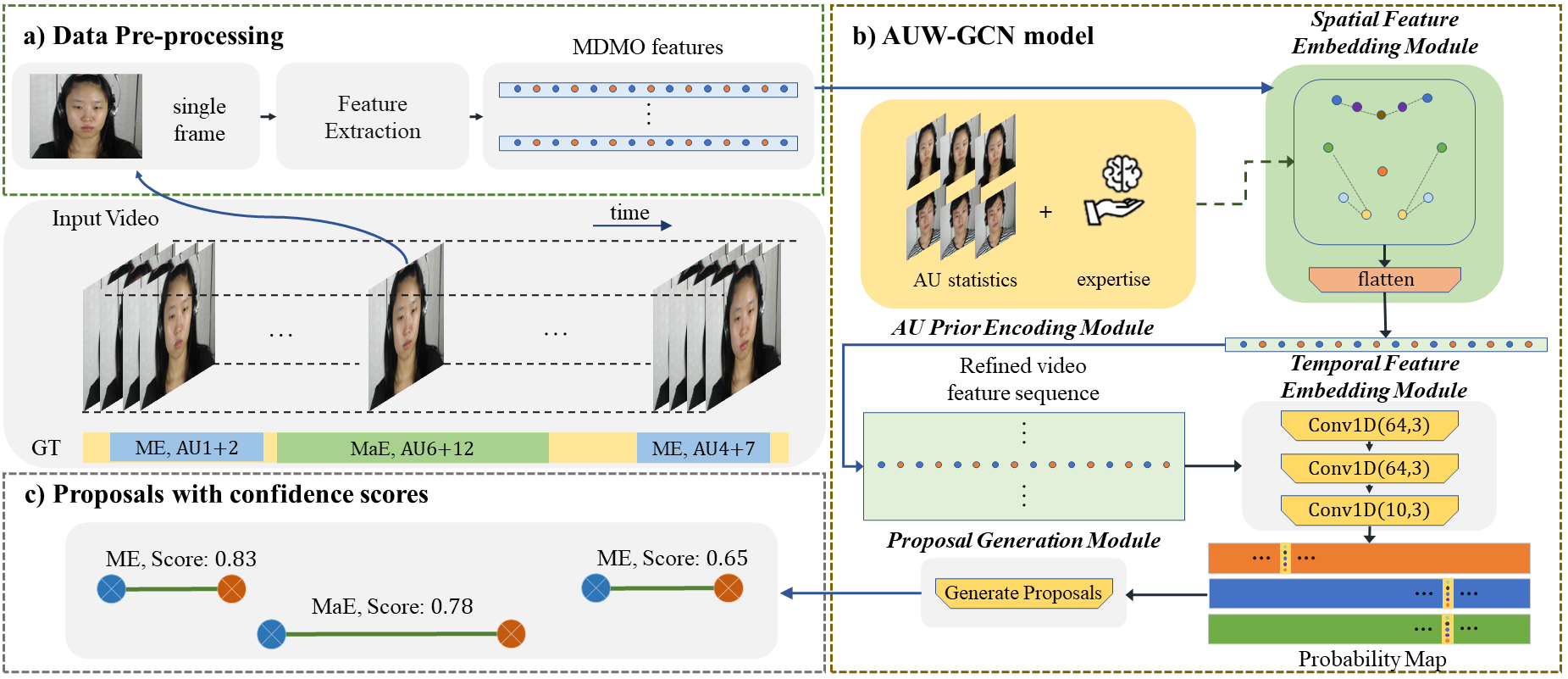}}
\end{minipage}
\end{center}
   \caption{Illustration of our framework. After feature extraction, we feed MDMO features into SFEM and TFEM to get probability maps, which are later used for proposal generation.}
\label{fig:framework}
\end{figure*}

\section{Related Works}
\subsection{ME Spotting}
Current methods for the ME Spotting task can be broadly divided into two categories: hand-crafted and deep-learning methods. The former entails hard-wired feature engineering and careful design of signal processing. Specifically, the procedure mainly consists of hand-crafted features design, feature difference analysis, and threshold strategy for determining the intervals of expressions~\cite{liong2022mtsn}. Typical feature descriptors used in these works include LBP~\cite{moilanen2014spotting}, optical-flow~\cite{he2020spotting, yuhong2021research, zhang2020spatio} and HOG~\cite{davison2018objective}.

With the development of deep learning, automatic feature engineering and end-to-end training have become a possibility. As a result, deep-learning-based methods are raised to a mainstream paradigm. Generally, researchers will choose a standard architecture, including CNNs and RNNs, and treat the task as a typical supervised problem. For example, Wang et al.~\cite{wang2021mesnet} used 2D-CNNs and 1D-CNNs to extract spatial and temporal features, respectively, with a regression network to refine the interval proposals. Yap et al.~\cite{yap20223d} adopted 3D-CNNs to extract spatial and temporal features simultaneously. Leng et al.~\cite{leng2022abpn} adapted BSN~\cite{lin2018bsn} for spotting intervals of ME and MaE flexibly, and we take inspiration from this work to design our network.

\subsection{Graph Convolutional Network}
By treating a human face as a graph with different ROIs as nodes, Liu et al.~\cite{liu2020relation} adopted GCN for AU detection. They built the graph with binary value through setting a threshold on the conditional probability of pairs of AUs. Similarly, Lo et al.~\cite{lo2020mer} used GCN for the ME recognition task. However, applying GCN in ME spotting has been left unexplored. We thus utilize the modelling capacity of GCN and further extend the approach in the field of MaE and ME spotting. 

\section{Methods}
Our methods involve feature extraction, the proposed AUW-GCN model, and optimization strategies. The overview of our approach is illustrated in Fig.\ref{fig:framework}.

\subsection{Feature Extraction}
Following ~\cite{leng2022abpn}, the feature extraction procedure includes facial alignment, landmarks detection, optical-flow calculation, and MDMO~\cite{liu2015main} feature extraction. 

Firstly, we use Retinaface ~\cite{deng2020retinaface} to locate the facial bounding box of the first frame in the video and align the subsequent frames to the first one. We denote this aligned video as $X_V=\{x_n\}_{n=1}^{l_v}\in \mathbb{R}^{H\times W\times l_v}$, where $H, W$ are the height and width of a single frame, respectively. $l_v$ represents the total frame numbers of the video. $x_n$ is the n-th frame in the video. After that, we adopt TV-L1 algorithm~\cite{zach2007duality} to compute coarse optical-flow features, denoted as $O=\{o_n\}_{n=1}^{l_v-1} \in \mathbb{R}^{H\times W \times (l_v-1) \times 2}$, where $o_n$ is the n-th optical-flow map calculated from $x_n$ and $x_{n+1}$. To refine the features, we follow the approach of ~\cite{liu2015main} and choose 12 ROIs and crop them out with facial landmarks detected by SAN~\cite{dong2018style}. Finally, we compute the finer-grained MDMO features as the model input, denoted as $F=\{f_n\} _{n=1}^{l_v-1} \in \mathbb{R}^{(l_v-1) \times N \times 2}$, where N=12 is the number of ROIs.

\begin{table}[!htbp]
\small 
\centering
\caption{The detailed architecture of AUW-GCN, following a  ``\textit{backbone-neck-head}'' design paradigm. RF is the equivalent size of the temporal receptive field. For the GCN layer, we denote the numbers of nodes and hidden dimensions in the form of (N, d).}
\resizebox{0.5\textwidth}{!}{
\begin{tabular}{c | c| c c c c c |c}
\hline
 \multicolumn{2}{c|}{layer} & kernel & stride & dim &RF & act. & output size\\ \hline
 
\multirow{2}{*}{\bf{Backbone}}
& GCN  &- &-  &(12,16) &- & relu & 16$\times$12$\times$T\\
\cline{3-8}
& flatten &- &- &- &- &- & 192$\times$T\\
\hline

\multirow{2}{*}{\bf{Neck}}
 & $conv1d_1$ & 3 &  1 &  64 &3 & relu & 64$\times$T\\
\cline{3-8}
 & $conv1d_2^*$ &  3 &  1 &  64 &7 & relu  & 64$\times$T\\

\hline
 \bf{Head} 
 & $conv1d_3^*$ & 3 & 1 & 10 &11 &-   & 10$\times$T\\
\hline
\end{tabular}
}

\label{table:arch_layer}
\begin{tablenotes}
    \item{*:} The last two 1D-convolution layers are dilated convolution with a dilation rate set to 2.
\end{tablenotes}
\end{table}

\subsection{AUW-GCN model}
As Table~\ref{table:arch_layer} shows, the whole architecture follows a standard detection design paradigm, i.e., backbone, neck, and head. The outputs of our model are preliminary proposals, which are then post-processed to get final spotting results. In this part, we introduce them in sequence, just as how the data flows.

\begin{figure}[h]
    \centering
    \includegraphics[width=0.5\textwidth]{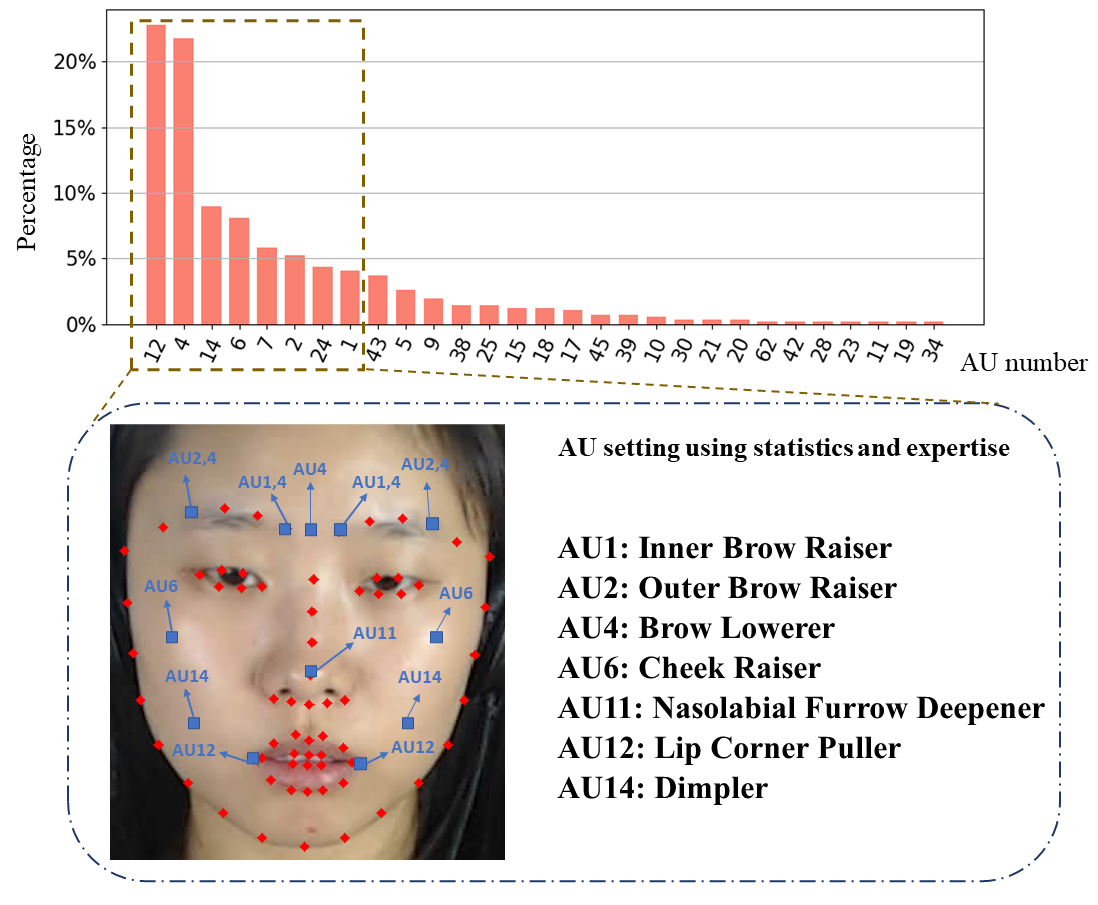}
    \caption{Illustration of AU selection and correspondence between AUs and ROIs.}
    \label{fig:AU_setting}
\end{figure}

\subsubsection{AU Prior Encoding Module}
\label{section:GT-definition}
To learn fine-grained relationships in small datasets, we resort to prior information and try to embed this information into GCN. Then naturally, the central problems are how to characterize the correlations of different ROIs and how to build the adjacency matrix. Consequently, we associate co-occurrences with correlations of AUs and map AUs to specific ROIs on human faces. In view of the critical observation that the ratio of AUs follows a long-tailed distribution and some field expertise, we choose 12 ROIs as Fig.~\ref{fig:AU_setting} shows. Then the remaining problem is how to quantify these relationships in the form of an adjacency matrix. Inspired by ~\cite{lo2020mer}, we devise a simple yet effective strategy for exploiting AU information in the dataset, which can be encoded as prior belief into our GCN. For more explicit demonstration, we denote the label set as $\Phi_g=\{\phi_n=(t_{s,n}, t_{ap,n}, t_{e, n}, \{U_n\})\}_{n=1}^{N_g}$, where $N_g$ is the total number of ground-truth labels, each made up of a quadruple, wherein $t_{s,n}, t_{ap,n}, t_{e, n}$ is the onset, apex, offset frame of the $n$-th ground-truth instance $\phi_n$, and $\{U_n\}$ is the set of AUs appearing in the instance. Formally, we construct the adjacency matrix $A^\prime$ in the following way:
\begin{equation}
\begin{aligned}
A^{\prime}_{ij} &= \sum_{\phi_k}\sum_{i,j}  \mathbbm{1}(i\in f(U_i),j \in f(U_j)), \quad U_i,U_j \in \{U_k\}.
\end{aligned}
\end{equation}
$f(\cdot)$ is a mapping function from a certain AU to a set containing all the corresponding facial regions as shown in Fig~\ref{fig:AU_setting}.
We normalize the matrix to get the final adjacency matrix $A$ to ensure training stability. The adjacency matrix formulated in this way is sparse and efficient but, at the same time, reserves the most representative information. 
\subsubsection{Spatial Feature Embedding Module}
To characterize the relationships between different parts of human faces and achieve finer spatial feature embedding, we choose GCN as a basic building block for the spatial feature embedding module (SFEM). GCN takes the feature representation matrix $X\in \mathbb{R}^{d\times N}$ and the adjacency matrix $A\in \mathbb{R}^{N\times N}$ as input, where d is the dimension of input features, and N is the number of nodes of the graph. We can stack L GCN layers one upon another. The outputs of the $l$-th GCN layer can be expressed as:
\begin{equation}
X^l=\sigma(AX^{l-1}W^{l-1}),
\end{equation}
where we choose ReLU as activation function~$\sigma$. $A$ is our prior-encoded adjacency matrix and $X^{l-1}, W^{l-1}$ are the feature embedding and learnable weight matrix of the $(l-1)\text{-th}$ layer respectively. The input of the module $X_0=F$ is the extracted MDMO feature, and the final output $X^L$ is the refined spatial feature embedding after graph convolution. 

\subsubsection{Temporal Feature Embedding Module}
This module operates temporal convolution on fine-grained spatial features $X^{L}$ from SFEM and outputs probability maps, which can be expressed as:
\begin{equation}
    P=\{p_t^s,p_t^{ap},p_t^e, p_t^{exp}\},
\end{equation}
where $p_t^s,p_t^{ap},p_t^e, p_t^{exp}$ 
 respectively represent the probability that each frame is classified as an onset, apex, offset, and expression frame. Here we define an expression frame as a frame inside a ground-truth interval. There are two main factors that affect our design choices, the size of the receptive field and the number of parameters. Given the data characteristics, our objective is to design, with a tight budget on model capacity, a model whose temporal receptive field is large enough to incorporate contextual information for mitigating uncertainty. Thus, in order to prevent over-fitting, we use dilated convolution in the last two layers of the module for a lighter architecture.

\normalem 
\begin{algorithm}[!ht]
\DontPrintSemicolon
  
  \KwInput{Probability sequences~$p_t^s,p_t^{ap},p_t^e$, threshold~$thr_{ap}$, search range~$k_{dis}$} 
  \KwOutput{Proposal set $\Phi_{p}=\{\phi_i=(\hat{t}_{s,i}, \hat{t}_{e,i}, sc_{p,i})\}_{i=1}^{N_p}$}

  $\Phi_{p} \gets \emptyset$\;
  $\Phi_p^{ap} \gets \emptyset$\;
  \ForEach{apex probability $p_i^{ap} \in p_t^{ap}$}{
      \If {$p_i^{ap} \geq thr_{ap}$} {
      $\Phi_p^{ap} \gets \Phi_p^{ap} \cup \{i\}$
    }
  }

  \ForEach{apex frame index $i \in \Phi_p^{ap}$}{
      $sc_{ap} \gets p_t^{ap}[i]$\;
      $\hat{t}_{s} \gets \mathop{\mathrm{argmax}}\limits_{i-k_{dis} \leq t^\prime_s \leq i - 1} {p_t^s[t^\prime_s]}$\;
      $sc_s \gets p_t^s[\hat{t}_s]$\;
      $\hat{t}_{e} \gets \mathop{\mathrm{argmax}}\limits_{i+1 \leq t^\prime_e \leq i + k_{dis}}    {p_t^e[t^\prime_e]}$\;
      $sc_e \gets p_t^e[\hat{t}_e]$\;
      $sc_p \gets sc_s \times sc_{ap} \times sc_e$\;
      $\Phi_{p} \gets \Phi_{p} \cup \{(\hat{t}_{s}, \hat{t}_{e}, sc_p)\}$\;
  }

  \KwRet{$\Phi_{p}$}\;
\caption{Generating the candidate proposal set from probability sequences.}
\label{alg:proposal}
\end{algorithm}
\ULforem 

\subsubsection{Proposal Generation Module}
After we get the probability sequences for the onset, apex, and offset frame, we generate candidate proposals as described in~Alg.~\ref{alg:proposal}.

\subsubsection{Post processing}
By further processing candidate proposals, we reduce highly-overlapping intervals and thus improve the quality of intervals spotted. The NMS~\cite{neubeck2006efficient} algorithm is adopted to filter out proposals whose scores are lower but are highly overlapped with confident ones, resulting in a final proposal set: $\Phi^\prime_{p}=\{\phi_n=(\hat{t}_{s,n}, \hat{t}_{e,n}, sc_{p,n})\}_{n=1}^{N^\prime_p}$, where $N^\prime_p$ is the number of final proposals.
\subsection{Optimization}
We convert the spotting task into a common classification problem. Specifically, we devise a binary classification task and a 3-class classification task for different types of frames. To cope with data imbalance, we choose Focal Loss~\cite{lin2017focal} as our basic loss function. Generally, the loss function can be expressed as:
\begin{equation}
\begin{aligned}
L(y, \hat{y})=-\frac{1}{l_w} & \sum_{i=1}^{l_w} \sum_{c=1}^{C}  [\alpha \cdot (1-y_{i,c})^\gamma \cdot y_{i,c} \cdot \log \hat{y}_{i,c} \\
+ & (1-\alpha)\cdot y_{i,c}^\gamma \cdot(1-y_{i,c}) \cdot \log (1-\hat{y}_{i,c})],
\end{aligned}
\end{equation}
where $C$ is the number of classes, $\hat{y}$ is the output probability of the model and $y$ is the ground-truth label. $\alpha, \gamma$ are hyper-parameters to balance positive-negative and easy-hard samples, respectively. Note that we use the sliding window technique to segment a long video into snippets, and $l_w$ is the window size.

\begin{table*}[h]
\tiny 
\centering
\caption{Spotting results on $\text{CAS(ME)}^2$ and SAMM-LV datasets in terms of F1-score}
\label{table_comparison_SOTA} 
\resizebox{\textwidth}{!}{
\begin{tabular}{|c|c|c|c|c|c|c|c|c|}
\hline

\multicolumn{2}{|c|}{\multirow{2}{*}{Methods}} & \multicolumn{3}{c|}{$\text{CAS(ME)}^2$} & \multicolumn{3}{c|}{SAMM Long Videos} & \multirow{2}{*}{Overall} \\ \cline{3-8}
\multicolumn{2}{|c|}{} & MaE & ME & Overall & MaE & ME & Overall & \\ \hline

\multirow{3}{*}{Hand-crafted methods} & MDMD~\cite{he2020spotting} &0.1196 &0.0082 &0.0376 &0.0629 &0.0364 &0.0445 &0.0445 \\ \cline{2-9}
& SP-FD~\cite{zhang2020spatio} &0.2131 &0.0547 &0.1403 &0.0725 &0.1331 &0.0999 &0.1243 \\ \cline{2-9}
& OF-FD~\cite{yuhong2021research} &0.3782 &\bf{0.1965} &0.3436 &0.4149 &\bf{0.2162} &0.3638 &0.3534 \\ \cline{3-8} \hline 

\multirow{6}{*}{Deep-learning methods}
& SOFTNet~\cite{liong2021shallow} &0.2410 &0.1173 &0.2022 &0.2169 &0.1520 &0.1881 &0.3006 \\ \cline{2-9}
& 3D-CNN~\cite{yap20223d} &0.2145 &0.0714 &0.1675 &0.1595 &0.0466 &0.1084 &- \\ \cline{2-9}
& Concat-CNN~\cite{yang2021facial} &0.2505 &0.0153 &0.2019 &0.3553 &0.1155 &0.2736 &0.2452 \\ \cline{2-9}
& LSSNet~\cite{yu2021lssnet} &0.3770 &0.0420 &0.3250 &0.2810 &0.1310 &0.2380 &0.2717 \\ \cline{2-9}
& MTSN~\cite{liong2022mtsn} &0.4104 &0.0808 &0.3620 &0.3459 &0.0878 &0.2867 &0.3191 \\ \cline{2-9}
& \bf{AUW-GCN (Ours)} &\bf{0.4235} &0.1538 &\bf{0.3834} &\bf{0.4293} &0.1984 &\bf{0.3728} &\bf{0.3771} \\ \hline

\end{tabular}
}
\end{table*}

\section{Experiments}
\subsection{Experimental Settings}
Following the protocol of MEGC2021~\cite{li2021fme}, we employ the Leave-One-Subject-Out (LOSO) cross-validation strategy for our experiments. 

\noindent
{\bf Datasets.}
We validate our methods and conduct experiments on two benchmark datasets:
{\bf $\text{CAS(ME)}^2$}~\cite{qu2017cas} dataset has 98 annotated videos from 22 subjects with 30 fps and 357 ground-truth instances, including 57 ME labels and 300 MaE labels; {\bf SAMM-LV}~\cite{yap2020samm} is a dataset of 224 long videos with 200fps recorded from 32 subjects. The dataset contains 159 ME samples and 343 MaE samples.  

\noindent
{\bf Implementation Details.}
The model is trained by Adam optimizer on both datasets for 100 epochs with a learning rate of 0.01. The thresholds for apex score and NMS post-processing are set to 0.4 and 0.5, respectively. 

\noindent
{\bf Evaluation Metrics.}
We follow the standard evaluation protocol used in the MEGC2021 spotting track. A proposal is considered a True-Positive (TP) if it satisfies the following:
\begin{equation}
    \frac{W_{proposal} \cap W_{GroundTruth}} {W_{proposal} \cup W_{GroundTruth}} \geq k_{IoU},
\end{equation}
where $k_{IoU}$ is set to 0.5 officially. Otherwise, the proposal counts as a False-Positive (FP). We calculate precision, recall, and F1-score for a more comprehensive comparison against other methods.

\subsection{Experimental Results}
\noindent
{\bf Overall Results.}
We report the performance of our methods on MEGC2021 benchmarks: $\text{CAS(ME)}^2$ and SAMM-LV, and comprehensively compare our results with hand-crafted methods and deep-learning methods by following~\cite{liong2022mtsn}. The results are listed in Table~\ref{table_comparison_SOTA}.
As the table shows, our methods outperform others consistently on both datasets, especially on SAMM-LV.
Our overall f1-score reaches 0.3653 on $\text{CAS(ME)}^2$ and 0.3706 on SAMM-LV. This is the first time that deep-learning methods beat SOTA hand-crafted method, OF-FD~\cite{yuhong2021research}. Though our performance on ME is lower, it should be noted that the latter requires laborious feature engineering and heavy tuning of parameters, while our methods can adaptively learn fine-grained feature embedding and exempt the chore of complicated tuning. Moreover, deep-learning methods excel in generalization, which is a promising prospect with the fast development of new and larger datasets.

\noindent
{\bf Ablation Studies.}
To further verify the effectiveness of our proposed methods, we conduct empirical studies on the components of our model based on the $\text{CAS(ME)}^2$ dataset.

\begin{table}[h]

\small
\centering
\caption{ Study of the effectiveness of SFEM and contribution of each component in the module, i.e., GCN and Prior encoding. Note that without prior, the GCN learns the adjacency matrix from data through training.}
\begin{tabular}{p{0.8cm}<{\centering}p{0.8cm}<{\centering}p{1.0cm}<{\centering}p{1.0cm}<{\centering}p{1.15cm}<{\centering}}
\toprule
					GCN 	& Prior 	& Precision	& Recall 	& F1-score	\\
 \hline
 		\XSolidBrush			 	& \XSolidBrush			& 0.3527	& 0.2885	& 0.3174	\\
 		\Checkmark  		&\XSolidBrush	& 0.3804 & 0.2941	& 0.3318	\\
 					\Checkmark	&\Checkmark	& {\bf 0.4000}	& {\bf 0.3361}& {\bf 0.3653}	\\
\bottomrule
\end{tabular}
\label{table_ablation_arch}
\end{table}

\begin{figure*}[!htb]

\begin{center}
\begin{minipage}[b]{\textwidth}
  \centering
  \centerline{\includegraphics[width=0.98\textwidth]{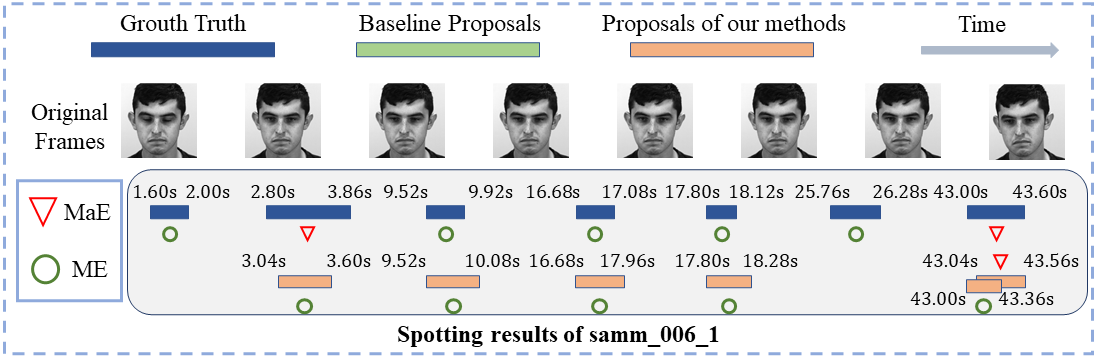}}
\end{minipage}
\end{center}
   \caption{Visualization analysis of spotting results. The example is from video samm\_006\_1 on SAMM-LV dataset.}
\label{fig:case_study}
\end{figure*}

To verify the effectiveness of introducing GCN for better spatial feature embedding, we compare the results of the model with and without GCN as Table~\ref{table_ablation_arch} shows. It is clear that introducing the SFEM improves all the metrics consistently. Specifically, the F1-score rises from 0.3174 to 0.3318. This improvement verifies the effectiveness of our proposed module in refining the feature embedding. Moreover, applying AU-prior encoding into the module brings another 10.0\% boost in F1-score and achieves a higher recall by 14.2\%. The enhancement validates the efficacy of our strategy for building an adjacency matrix from prior belief. This can be partially explained by the fact that given a small dataset, learning relationships between graph node embedding could be pretty tricky, and may even fall into over-fitting. In contrast, by injecting specific knowledge into the model, the network can learn correlations between nodes more efficiently. 

\begin{table}[h]
\small
\centering
\caption{The results of ablation study on model design choices. NL = number of GCN layers. ND = hidden dimension of GCN.}
\begin{tabular}{p{1.0cm}<{\centering}|p{0.5cm}<{\centering}p{0.5cm}<{\centering}p{1.0cm}<{\centering}p{1.0cm}<{\centering}p{1.15cm}<{\centering}}
\toprule
			Setting		& NL 	& ND 	& Precision	& Recall 	& F1-score	\\
 \hline
 	a				& 1	& 16 			& {\bf 0.4000}	& {\bf 0.3361}& {\bf 0.3653}	\\
        b		&1  		&32	& 0.1025 & 0.1625	& 0.1257	\\
 	c				&2	&16	& 0.3869	& 0.2969 & 0.3360	\\
        d      &2          &32         &0.0850          &0.1092       &0.0956 \\
\bottomrule
\end{tabular}
\label{table_ablation_GCN_param}

\end{table}

We also investigate the effects of different model capacities by varying the number of layers and hidden dimensions as listed in Table~\ref{table_ablation_GCN_param}. Compared with setting (a), setting (b) doubles the hidden dimension and results in severe performance degradation. Moreover, increasing both the number of layers and hidden dimension (setting (d)) causes a drop in the F1-score from 0.3653 to 0.0956. In contrast, only adjusting the number of GCN layers (setting (c)) brings a slight degradation. These results overall suggest that larger capacity is inappropriate for small datasets, and our model design is better suited.

\subsection{Case Study}
For an intuitive illustration, we show a qualitative example in Fig.~\ref{fig:case_study}. As the figure shows, our model can generate proposals with high recall for both MaEs and MEs owing to the temporal modeling capacity of TFEM. The result suggests that leveraging contextual information can help the model locate more accurate and complete intervals. 

\section{Conclusions}
This paper proposed a graph-based network for spotting MaEs and MEs, dubbed AUW-GCN. To assist the model in learning relationships between different regions of human faces, a strategy was devised for injecting AU-prior information through careful design of the adjacency matrix of the GCN module. Moreover, comprehensive experiments on two benchmark datasets demonstrated our methods' superior effectiveness and generalization ability. 

\clearpage

\bibliographystyle{IEEEbib}
\bibliography{icme2023template}

\end{document}